\documentclass[journal,twoside,web]{ieeecolor}
\usepackage{generic}
\usepackage{cite}
\usepackage{amsmath,amssymb,amsfonts}
\usepackage{algorithmic}
\usepackage{graphicx}
\usepackage{algorithm,algorithmic}
\usepackage{hyperref}
\hypersetup{hidelinks=true}
\usepackage{textcomp}
\usepackage{amsmath, xurl, pdfpages, caption, subcaption, makecell, fancyhdr}
\usepackage{graphics,amssymb,epsfig,color}
\usepackage{comment}
\usepackage{fixltx2e}
\usepackage{placeins}
\usepackage{authblk}
\usepackage{comment}
\usepackage{array}
\usepackage{dblfloatfix}
\newcolumntype{P}[1]{>{\centering\arraybackslash}p{#1}}

\usepackage[british]{babel}
\usepackage{hhline}
\usepackage{multirow}

\usepackage{amsmath}
\def\BibTeX{{\rm B\kern-.05em{\sc i\kern-.025em b}\kern-.08em
    T\kern-.1667em\lower.7ex\hbox{E}\kern-.125emX}}
\begin{document}

\title{SleepPPG-Net2: Deep learning generalization for sleep staging from photoplethysmography}
\author[1, 2]{Shirel Attia}
\author[3]{Revital Shani Hershkovich}
\author[3]{Alissa Tabakhov}
\author[2]{Angeleene Ang}
\author[2]{Sharon Haimov}
\author[3]{Riva Tauman}
\author[2]{Joachim A. Behar}

\affil[1]{The Faculty of Data and Decision Sciences, Technion, Israel Institute of Technology, Haifa, Israel}
\affil[2]{Faculty of Biomedical Engineering, Technion, Israel Institute of Technology, Haifa, Israel}
\affil[3]{The Institute for Sleep Medecine, Ichilov, Tel-Aviv, Israel }

\maketitle

\begin{abstract}
Background: Sleep staging is a fundamental component in the diagnosis of sleep disorders and the management of sleep health. Traditionally, this analysis is conducted in clinical settings and involves a time-consuming scoring procedure. Recent data-driven algorithms for sleep staging, using the photoplethysmogram (PPG) time series, have shown high performance on local test sets but lower performance on external datasets due to data drift. Methods: This study aimed to develop a generalizable deep learning model for the task of four class (wake, light, deep, and rapid eye movement (REM)) sleep staging  from raw PPG physiological time-series. Six sleep datasets, totaling 2,574 patients recordings, were used. In order to create a more generalizable representation, we developed and evaluated a deep learning model called SleepPPG-Net2, which employs a multi-source domain training approach.SleepPPG-Net2 was benchmarked against two state-of-the-art models. Results: SleepPPG-Net2 showed consistently higher performance over benchmark approaches, with generalization performance (Cohen’s kappa) improving by up to $19\%$. Performance disparities were observed in relation to age, sex, and sleep apnea severity. Conclusion: SleepPPG-Net2 sets a new standard for staging sleep from raw PPG time-series.
\end{abstract}

\IEEEpeerreviewmaketitle

\begin{figure*}[htb!]
  \centering
\includegraphics[width=\textwidth]{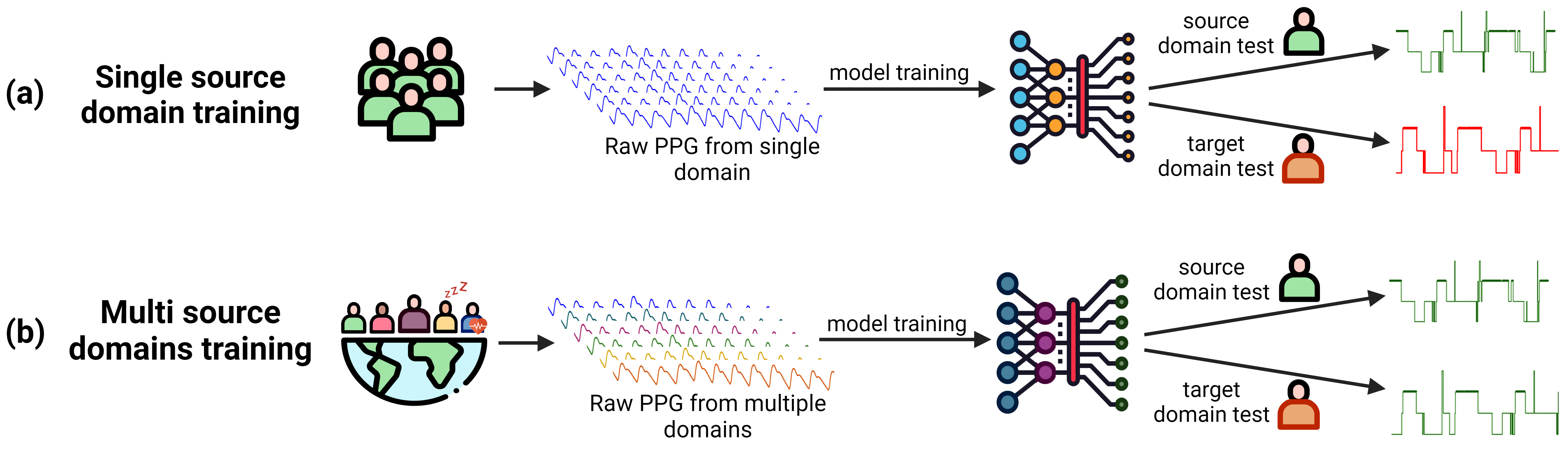}
    \caption{Research overview. Panel (a) presents a classic strategy that involves the training of a sleep staging classification model using a single dataset (source domain), which typically exhibits decreased performance on external datasets (target domains). Panel (b) introduces our multi-source training strategy, which utilizes multiple datasets during training to develop a more generalizable representation, thereby enhancing performance when the model is evaluated on an external dataset (target domain).}
    \label{fig:abstract}
\end{figure*}


\begin{figure*}[htb!]
  \centering
\includegraphics[width=\textwidth]{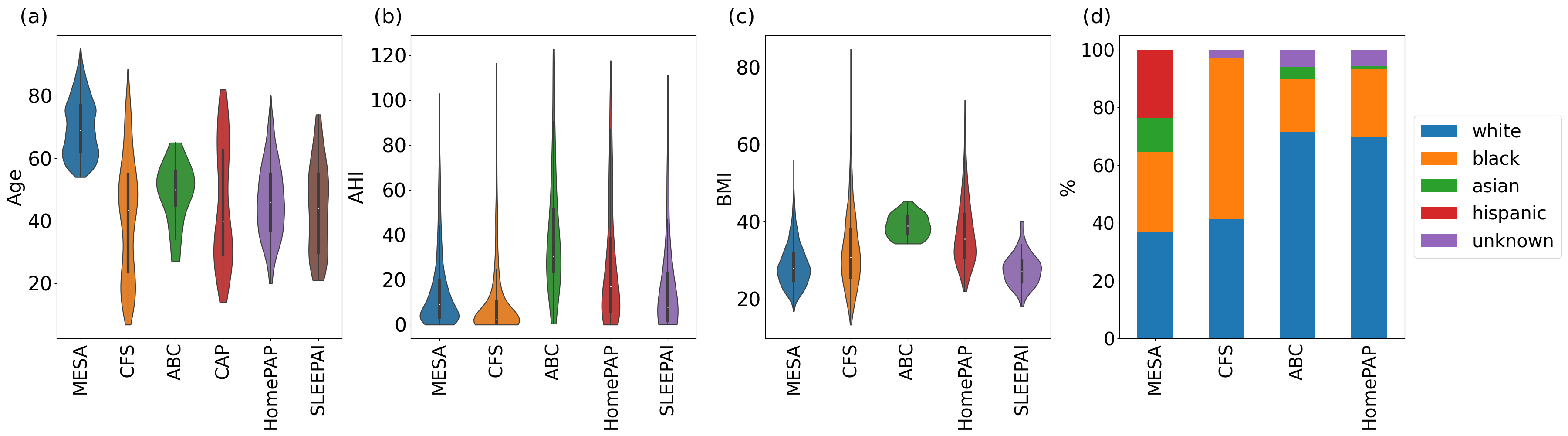}
    \caption{Data distribution presented in violin plots for (a) age, (b) apnea hypopnea index (AHI) and (c) body mass index (BMI) and bar plot for (d) ethnicity. The AHI, BMI and ethnicity variables were not available for the CAP dataset and ethnicity was not available for the SLEEPAI dataset.}
    \label{fig:distribution_violin}
\end{figure*}

\section{Introduction}

Sleep plays a critical role in health and well-being. During sleep, the brain engages in complex activities characterized by dynamic and cyclic patterns. Clinicians have systematically classified sleep neural patterns into distinct stages, facilitating the evaluation and analysis of sleep and enabling the detection of sleep disorders. Physiological signals used to stage sleep include the electroencephalogram, electrooculogram, and electromyogram. These, together with respiratory signals, electrocardiogram (ECG) and video, are conventionally recorded during polysomnography (PSG), the gold standard sleep analysis tool, which is conducted in a clinical setting. Traditionally and to this day, the classification of sleep stages is performed manually by trained technicians, according to the American Academy of Sleep Medicine (AASM) guidelines \cite{berry_aasm_2017}. 

Photoplethysmography (PPG)-based sleep staging is becoming a popular alternative to in-lab PSG. PPG is a noninvasive method that detects volumetric changes in the microvascular tissue bed and serves as an important sensing technology incorporated into contemporary wearable devices, including smartwatches and fitness trackers ~\cite{charlton_2023_2023}. It is already widely used to monitor heart rate and blood oxygen saturation, and its application in sleep analysis has garnered substantial interest.

A considerable amount of research has focused on developing methods to enhance sleep staging accuracy using PPG. PPG-based sleep staging has evolved from methods using rhythm and morphological features, followed by thresholding or machine learning-based classification, and has given rise to a wide variety of deep learning-based algorithms ~\cite{kotzen_sleepppg-net_2023, sridhar_deep_2020, radha_deep_2021, huttunen_assessment_2021, habib_performance_2023, zhao_multi-class_2021, li_transfer_2021, korkalainen_deep_2020, walch_sleep_2019}. These algorithms can be broadly categorized by their use of beat-to-beat intervals~\cite{sridhar_deep_2020, zhao_multi-class_2021, li_transfer_2021} versus raw PPG data ~\cite{kotzen_sleepppg-net_2023, radha_deep_2021, habib_performance_2023, huttunen_assessment_2021, korkalainen_deep_2020, walch_sleep_2019} as input to the neural network algorithm. 

They are typically assessed against the gold-standard manual sleep staging, with Cohen's kappa employed to gauge inter-rater reliability. The majority of research report kappa performance on a local test set within the range of 0.62-0.77 \cite{radha_deep_2021, huttunen_assessment_2021, habib_performance_2023, zhao_multi-class_2021, wulterkens_it_2021}. Habib et al. \cite{habib_performance_2023} reported a kappa value of k=0.77 for a four-class classification task; however, the study was limited by a small dataset size (n=10) and employed a leave-one-subject-out cross-validation technique. A study conducted by Zhao et al. \cite{zhao_multi-class_2021} obtained k=0.69 on the cyclic alternating pattern (CAP) sleep dataset  \cite{goldberger_physiobank_2000, terzano_atlas_2001}, utilizing a leave-one-subject-out cross-validation method. Of note, data of only 27 out of 63 subjects was used. Few studies \cite{kotzen_sleepppg-net_2023, walch_sleep_2019}  have evaluated the generalization performance of their algorithms on target domains. Walch et al. \cite{walch_sleep_2019} evaluated an algorithm on the unseen multi-ethnic study of atherosclerosis (MESA) dataset but only on a limited portion of the available data, specifically 188 of the 2,052 subjects used in this study. In our previous work, we developed SleepPPG-Net \cite{kotzen_sleepppg-net_2023}, a deep learning model which takes the raw PPG data as input. SleepPPG-Net demonstrated state-of-the-art performance on the local test set \cite{charlton_2023_2023} but performance was reduced when evaluated on an independent external test set.

The challenge of generalization performance in physiological time-series analysis is particularly daunting due to the inherent complexity and high variability of human physiology.  Although high performance has been reported for deep learning models processing physiological signals~\cite{phan_seqsleepnet_2019,levy_deep_2023,ribeiro_automatic_2020}, the models tend to generalize poorly or moderately on external datasets ~\cite{kotzen_sleepppg-net_2023, levy_deep_2023, ballas_towards_2024}. This discrepancy is not just a technical issue but a significant barrier in translational research, where the ultimate goal is to apply these algorithms in clinical practice. Therefore, there is a pressing need for the development of generalizable models that not only perform well with initial datasets, but also ensure efficacy when deployed in a variety of health settings and population samples~\cite{behar_generalization_2023}.


This research seeks to expand the applicability of SleepPPG-Net \cite{kotzen_sleepppg-net_2023} by improving its generalization performance. In particular, by incorporating a multisource-domain training approach, we enhance the representation of the deep learning model through the inclusion of a diverse population sample encompassing a broad spectrum of demographic and physiological characteristics (Figure \ref{fig:abstract}). The intuition is that when training a model on a single dataset, it may overfit its specific domain distribution, while leveraging multiple datasets may enable learning of a more common representation and avoid learning shortcut features, thereby avoiding overfitting.

This paper starts with an introduction to the datasets used for the experiments. Then, we present the architecture and strategy used to train SleepPPG-Net2 and create a generalizable representation of the raw underlying PPG data for sleep-staging. Finally, a set of experiments is presented to evaluate the performance of the model in datasets with distribution shifts. The results are provided in the source domain test set and five independent target domain data sets, totaling 2,574 patients. This research makes the following key contributions:

\begin{itemize}
    \item The development of SleepPPG-Net2 and benchmark against state-of-the-art benchmarks: SleepPPG-Net \cite{kotzen_sleepppg-net_2023} and DTS \cite{sridhar_deep_2020}; 
    \item Evaluation of generalization performance on five independent data sets; 
    \item A quantitative error analysis pinpointing the main sources of classification errors.
\end{itemize}

The generalization performance was assessed in accordance with Level one described in Behar et. al. ~\cite{behar_generalization_2023}, i.e., limiting the effects of data drift or shortcut features by external validation on multiple retrospective datasets.

\begin{table}[tb!]
\centering
  \renewcommand{\arraystretch}{1.2}
\begin{tabular}{p{1.4cm}  p{0.7cm} p{0.6cm} p{0.6cm} p{0.8cm} p{0.8cm} p{1cm} }
    Dataset &  Number & Region & Male(\%) & fs(Hz)  & Test & Timeframe\\\hline 
    MESA\cite{zhang_national_2018}  & 2054 & USA & 46.5 & 256  & Type 1 & 2000–2002\\
    CFS\cite{redline_familial_1995}& 259 & USA & 40.0 & 128  & Type 2 & 2001–2006\\
    ABC\cite{bakker_gastric_2018} & 49 & USA & 55.3 &256 & Type 1 & 2011–2014 \\
    HomePAP\cite{rosen_multisite_2012}& 118 & USA & 54.2 & 25-256 & Type 1 & 2008–2010 \\
    CAP\cite{terzano_atlas_2001}& 63 & Italy & 63.0 & 128 & Type 1 & 2001 \\
    SLEEPAI  & 38 & Israel & 52.2 & 75  & Type 1 & 2023–2024\\
    \hline 
\end{tabular}
\caption{Datasets used for the experiments.  Number: number of individual each with one PSG. fs: PPG sampling frequency. The oximeter brand used for all the datasets was Nonin and unknown for HomePAP and CAP.}
\label{tab:summary_datasets}
\end{table}



\section{Datasets} \label{MATERIALS}
This study was approved by the Institutional Review Board of the Technion-IIT Rappaport Faculty of Medicine (62-2019). We used the raw PPG signal with metadata and the sleep staging scoring of PSG as a reference. Table I presents summary statistics for the six datasets used. PSG records with missing PPG data or technical errors and of individuals under 18 years of age were excluded. 

\subsection{Datasets}
\subsubsection{MESA}
The multi-ethnic study of atherosclerosis (MESA) dataset \cite{chen_racialethnic_2015, zhang_national_2018} consists of in-home PSG conducted with the Compumedics Somte System (Compumedics Ltd., Abbotsford, Australia). The PSG data were collected from 2,056 participants in four US communities. Overnight recordings were sent to the Brigham and Women's Hospital centralized reading center, where trained technicians scored the data according to the AASM rules. The PPG signal was captured using a Nonin 8000 sensor at a frequency of 256 Hz \ref{tab:summary_datasets}. Two patients were excluded due to missing data. The dataset represents a generally older age range compared to other datasets (Figure\ref{fig:distribution_violin}),  but a diversity of ethnicities including white, black / African American, Asian, and Hispanic (Table \ref{tb:summary_table_results}). The MESA dataset is divided into two parts: MESA-train, which comprises 90\% of the dataset and MESA-test, which includes the remaining 10\% of the dataset. This train-test split was carried out while stratifying per patient and age as described by Kotzen et al. \cite{kotzen_sleepppg-net_2023}.

\subsubsection{CFS}
The Cleveland family study (CFS)  \cite{chen_racialethnic_2015, redline_familial_1995}  includes individuals from 361 families in various communities, observed up to four times over 16 years. For this research, we had access to 320 PSG recordings from the fifth visit. We selected only adult participants (aged over 18 years), which excluded 61 young patients, resulting in  259 PSG. This visit included an overnight, in-lab 14-channel PSG using the Compumedics E-Series System (Abbotsford, Australia). The night recordings were sent to the centralized reading center at Brigham and Women's Hospital, and data were scored by trained technicians who used Rechtshaffen and Kales (R\&K) criteria for sleep staging and AASM rules for arousals. The PPG signal was recorded using a Nonin 8000 sensor at a frequency of 128 Hz (Table \ref{tb:summary_table_results}). The CFS dataset is unique in that it includes participants in a younger age range (Figure \ref{fig:distribution_violin}).

\subsubsection{ABC}
The apnea, bariatric surgery, and continuous positive airway pressure (CPAP) (ABC)  \cite{chen_racialethnic_2015, bakker_gastric_2018} dataset consists of in-lab PSG records collected using the Compumedics E-Series (Abbotsford, Victoria Australia). The primary objective of this dataset was to compare the efficacy of bariatric surgery versus CPAP therapy combined with weight loss counseling for treatment of 49 patients with class II obesity and severe obstructive sleep apnea (OSA). The PSG data including sleep stages and events were manually scored according to AASM guidelines. This dataset includes the same patients who had three full PSG assessments during the study period. For our purposes, we only used the baseline overnight in-lab PSG data. The PPG signal was recorded using a Nonin 8000 sensor at a frequency of 256 Hz. The ABC dataset typically showed higher AHI and BMI compared to other datasets (\ref{fig:distribution_violin}). 

\subsubsection{HomePAP}
The home positive airway pressure (HomePAP) \cite{chen_racialethnic_2015, rosen_multisite_2012} study was a multisite, non-blinded, randomized controlled trial performed in seven AASM-accredited academic sleep centers. It involved 373 patients selected due to their high probability of having moderate to severe OSA, determined by a clinical algorithm. In our study, we focused exclusively on the full in-lab PSG which amounted to 121 PSG recordings. However, 3 recordings had to be excluded due to the absence of the PPG signal, leaving us with 118 PSG recordings. Since the study was conducted across seven different clinical field sites, the oximeter and recording frequency varied for each patient (between 25 and 256 Hz, Table \ref{tb:summary_table_results}). The data scoring was centralized and conducted manually.

\subsubsection{CAP}
The cyclic alternating pattern (CAP) sleep dataset \cite{goldberger_physiobank_2000, terzano_atlas_2001} is a collection of 108 PSG recordings of CAP patients contributed by the Sleep Disorders Center of the Ospedale Maggiore of Parma, Italy. The PPG signal was recorded with a sampling frequency of 128 Hz. CAP, is a feature/component of non-rapid eye movement (NREM) microstructure that has been shown to signify sleep instability and sleep disturbances. CAP, while a physiological occurrence, is also an indicator of sleep instability and is associated with various sleep-related disorders such as sleep-disordered breathing, insomnia, sleep movement disorders (such as periodic leg movements (PLM) and restless leg syndrome (RLS)), parasomnias (including REM behavior disorder (RBD)) and neurological conditions (such as nocturnal frontal lobe epilepsy (NFLE) and narcolepsy). A total of 39 patients had missing channels and 6 other patients were excluded due to poor signal quality, resulting in 63 patients remaining for this dataset. The PSG records were manually scored using the Rechtshaffen and Kales (R\&K) criteria. 

\subsubsection{SLEEPAI}
The SLEEPAI dataset includes 50 PSG recordings from both OSA and non-OSA patients collected at the Sieratzki-Sagol Institute for Sleep Medicine of the Tel-Aviv Sourasky Medical Center, Israel. In addition to standard PSG recording, patients were also monitored using a wrist-worn oximeter during their PSG night at the sleep center. The dataset incorporates both the PPG signals from the wrist-worn oximeter and the corresponding scoring from the PSG recorded synchronously. The PPG signal was captured at 75 Hz (Table \ref{tb:summary_table_results}) and PSG were manually scored using the AASM guidelines (version 2.6). The study was approved by the Tel Aviv Sourasky Medical Center (0512-22) and informed consents were obtained from all participants.

\subsection{Labels harmonization}
The AASM guideline was used as reference for sleep staging definition. These guidelines categorize sleep stages into wake, REM and three NREM stages, namely N1, N2, and N3. In this study, we used a 4-class sleep staging system that includes wake, light sleep (N1/N2), deep sleep (N3) and REM. It is important to note that in the older Rechtschaffen and Kales (R\&K) guidelines, used by CFS, CAP, an additional NREM stage, S4 was included. In aligning the R\&K stages with our 4-class system, we included S4 as part of the deep sleep class. Therefore, our mapping consists of wake, light sleep (S1/S2), deep sleep (S3/S4), and REM.

\section{Methods} \label{Methods}

\subsection{Benchmarks}

\subsubsection{BM-DTS model}.
Sridhar et al. \cite{sridhar_deep_2020} developed a deep learning model that takes as input the instantaneous pulse rate (IPR), i.e. a time series derived from the interbeat intervals (IBIs) computed from ECGs. The benchmark Derived Time Series (BM-DTS) model architecture is based on this original work by Sridhar et al. \cite{sridhar_deep_2020}, with some minor modifications, as described in detail by Kotzen et al. \cite{kotzen_sleepppg-net_2023}. The BM-DTS model architecture takes as input the extracted continuous IPR time-series and is composed of three time-distributed residual convolutions and a time-distributed deep neural network. The DTS method hinges on the precise identification of PPG peaks. For this purpose, we used the validated pyPPG ~\cite{goda_pyppg_2023} PPG peak detector. BM-DTS was trained on MESA-train and results are provided for MESA-test and all target domains.


\subsubsection{SleepPPG-Net}
The following preprocessing treatment was performed on the raw PPG data as originally described by Kotzen et al. \cite{kotzen_sleepppg-net_2023}. A low-pass filter was used to eliminate high-frequency noise and prevent aliasing during the downsampling process. This filter was a zero phase 8th order low-pass Chebyshev Type II filter, with an 8Hz cutoff frequency and a 40dB stopband attenuation. The filtered PPG signal was downsampled using linear interpolation at a frequency of 34.13Hz. The PPG was then clipped within three standard deviations of the mean, and the data was standardized. The SleepPPG-Net \cite{kotzen_sleepppg-net_2023} architecture consists of a residual convolutional neural network for automatic feature extraction and a temporal convolutional neural network to capture long-range contextual information. In this work, SleepPPG-Net was pre-trained using the ECG SHHS dataset \cite{chen_racialethnic_2015, quan_sleep_1997} and then trained on the PPG of the MESA-train dataset (source domain) due to its large size. The performance of SleepPPG-Net is then reported for MESA-test as well as for all target domains. Model training was performed on a single GPU A100 (80Gb) using TensorFlow framework 2.8 and following an hyperparameter tuning procedure using Bayesian search over 100 iterations.


\subsection{SleepPPG-Net2}
In order to develop a generalizable deep learning model, some additions were added to the SleepPPG-Net architecture as well as training procedure. First, a domain-shifts-with-uncertainty (DSU) layer ~\cite{li_uncertainty_2022} with a probability of 0.5 was added to the architecture of SleepPPG-Net \cite{kotzen_sleepppg-net_2023} after the Dilated 1D-Convolution loop, followed by a Batch-Normalization layer and a Dropout layer. The DSU layer enhances the network's ability to generalize by capturing the uncertainty related to domain shifts with synthesized feature statistics during training. The model was pre-trained on the ECG SHHS dataset and then fine-tuned on a joint set of multiple-source datasets (four out of five), while evaluating generalization performance on the single left-out target domain. This training procedure was repeated five times in order to report performance on the one left-out target domain. This model is denoted SleepPPG-Net2.



\begin{table*}[htb!]
  \centering
  \renewcommand{\arraystretch}{1.2}
  \begin{tabular}{p{1.8cm} |p{0.4cm} p{0.4cm} p{0.4cm} | p{0.4cm} p{0.4cm} p{0.4cm}  |p{0.4cm} p{0.4cm} p{0.4cm} | p{0.4cm} p{0.3cm} p{0.4cm} | p{0.4cm} p{0.4cm} p{0.4cm} | p{0.4cm} p{0.4cm} p{0.4cm}}
    \hline
    \multirow{2}{3cm}
    {\textbf{Model}} 
    & \multicolumn{3}{c}{\textbf{MESA}} 
    & \multicolumn{3}{c}{\textbf{CFS}} 
    & \multicolumn{3}{c}{\textbf{ABC}} 
    & \multicolumn{3}{c}{\textbf{HomePAP}} 
    & \multicolumn{3}{c}{\textbf{CAP}} 
    & \multicolumn{3}{c}{\textbf{SLEEPAI}}\\ 
    \cline{2-19} 
    & \textbf{$\kappa_{p}$} & $\kappa_{c}$ & \textbf{$Acc$} & \textbf{$\kappa_{p}$} & $\kappa_{c}$ &  \textbf{$Acc$} & \textbf{$\kappa_{p}$} & $\kappa_{c}$ &  \textbf{$Acc$} & \textbf{$\kappa_{p}$} & $\kappa_{c}$ &  \textbf{$Acc$} & \textbf{$\kappa_{p}$} & $\kappa_{c}$ &  \textbf{$Acc$}& \textbf{$\kappa_{p}$} & $\kappa_{c}$ &  \textbf{$Acc$}\\
    \hline 
    DTS  & 0.65 & 0.64 & 0.77  & 0.63& 0.61 & 0.75  & 0.60 & 0.59 & 0.75  & 0.52& 0.49 & 0.68  & 0.35 & 0.39 & 0.57 & 0.53 & 0.55 & 0.73  \\ 
    SleepPPG-Net  & 0.74 & \textbf{0.74} & 0.83 & 0.72 & 0.71 & 0.81 & 0.67 & 0.66 & 0.79  & 0.63 & 0.64 & 0.77 & 0.48 & 0.49 & 0.64 & 0.61 & 0.62 & 0.77\\ 
    SleepPPG-Net2  & \textbf{0.75} & \textbf{0.74} & \textbf{0.84} & \textbf{0.74} & \textbf{0.73} & \textbf{0.82} & \textbf{0.69} & \textbf{0.68} & \textbf{0.80} &  \textbf{0.69} & \textbf{0.68} & \textbf{0.79} &  \textbf{0.57} & \textbf{0.54} & \textbf{0.67} &  \textbf{0.66} & \textbf{0.67} & \textbf{0.79} \\
    \hline
  \end{tabular}
  \caption{Performance for four-class classification, Median kappa ($\kappa_{p}$), overall kappa ($\kappa_{c}$) and accuracy ($Acc$). MESA is the source domain and the other datasets are target domains. }
  \label{tb:summary_table_results}
\end{table*}


\begin{figure*}[htb!]
  \centering
    \includegraphics[width=0.9\textwidth]{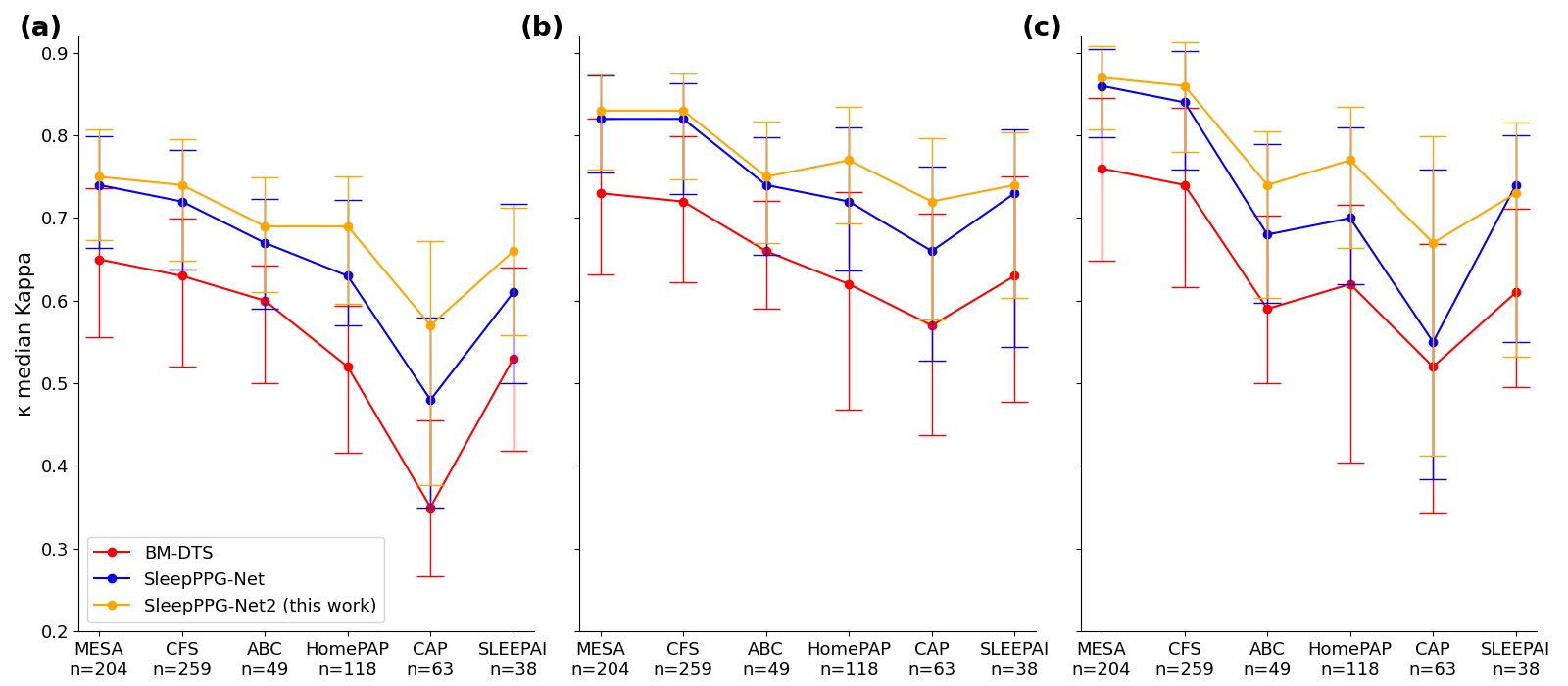}
    \caption{Median Kappa performance for (a) four-class (Wake, Deep, Light, REM), (b) three-class (Wake, NREM, REM) and (c) two-class (Wake, Sleep) sleep stage classification. The median performance is plotted for each test set.  The confidence interval for Kappa is provided as interquartiles (Q1-Q3).}
    \label{fig:res}
\end{figure*}

\begin{figure*}[htb!]
  \centering
    \includegraphics[width=0.9\textwidth]{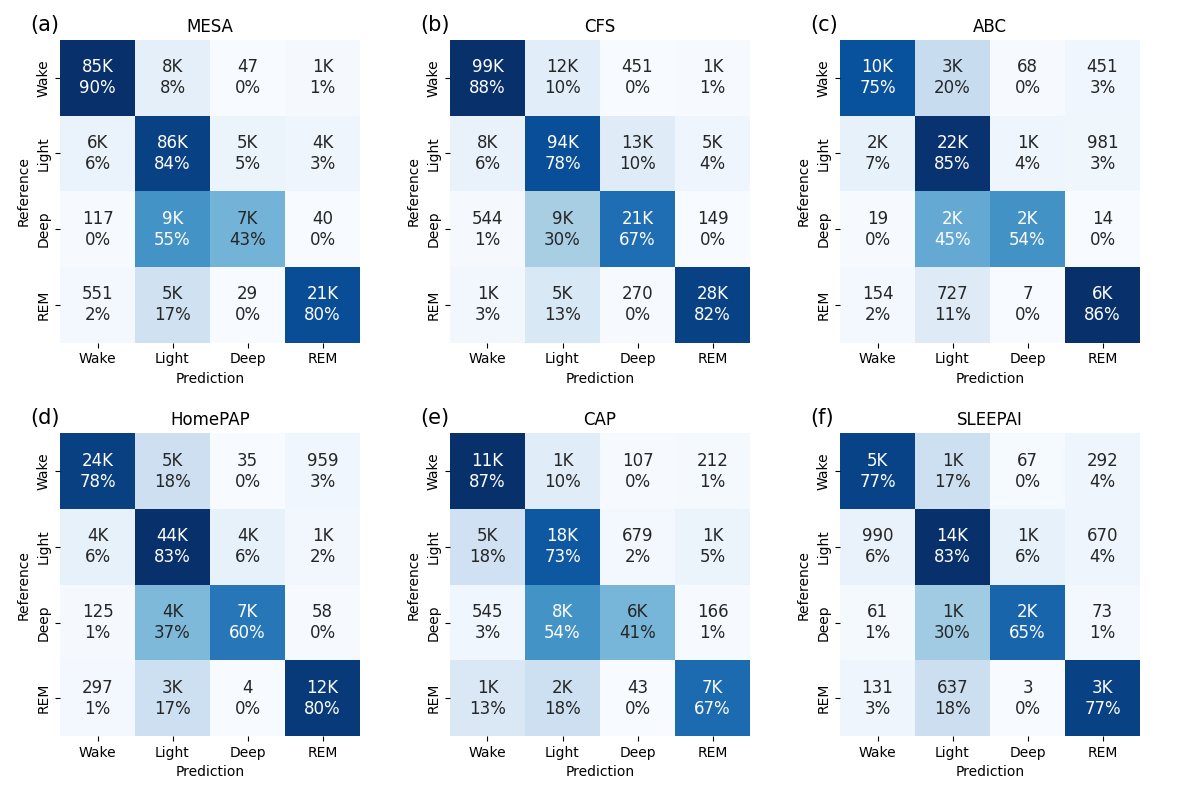}
    \caption{Confusion matrix for SleepPPG-Net2 (four-class). Source domain test set (a) and target domains (b-f).}
    \label{fig:cf}
\end{figure*}


\begin{figure*}[b!]
  \centering
    \includegraphics[width=\textwidth]{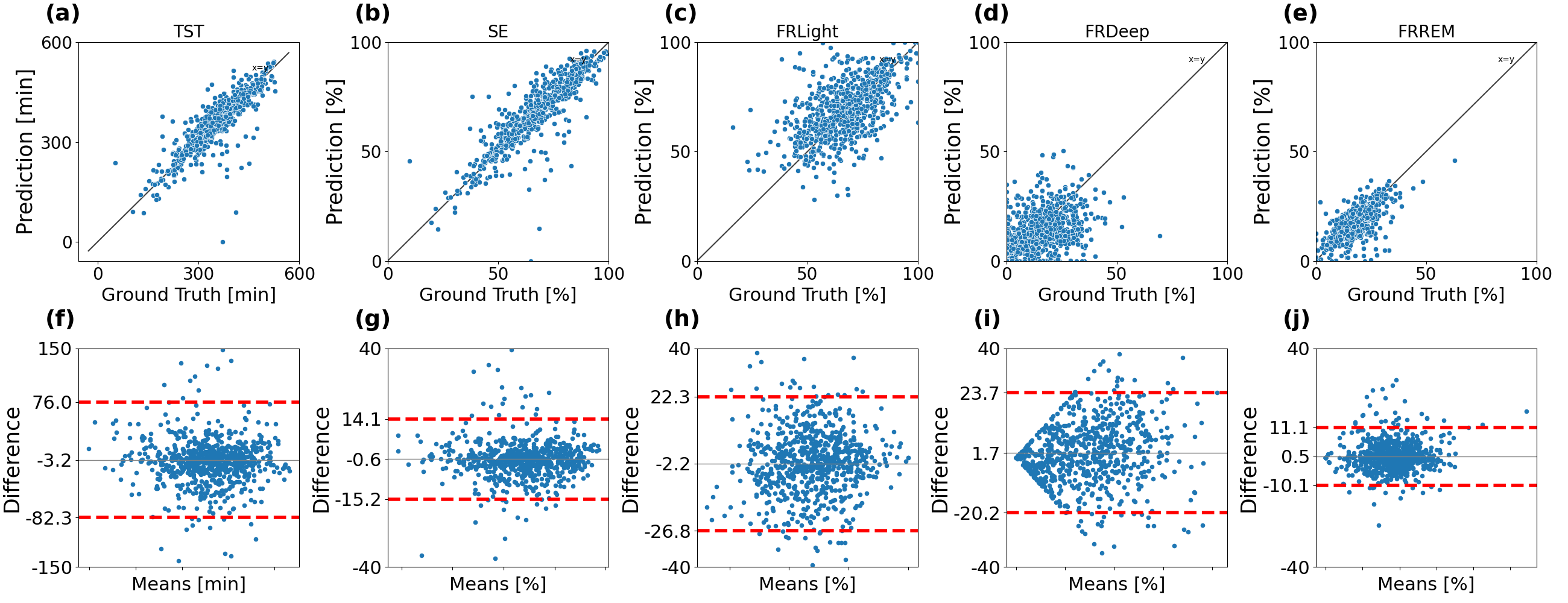}
    \caption{Scatter and Bland-Altman plots of the sleep measures. (a–e): Scatter plots of the ground truth and predicated sleep measures for all external datasets combined. The black line represents the equation y = x. (f–j): Bland-Altman plots comparing ground truth vs. estimated sleep measures for all external datasets combined. The error lines in red are positioned at ± 1.96 the standard deviation. From left to right the sleep measures are: (a, f)- TST in minutes, (b, g)- SE in percentage, (c, h)- FRLight in percentage, (d, i)- FRDeep in percentage, (e, j)- FRREM in percentage.}
    \label{fig:correlation}
\end{figure*}


\subsection{Performance assessment}

\subsubsection{Performance statistics}
The models output a probability prediction for each of the four sleep stages for each 30-sec sleep window. The probabilities were converted into predictions by selecting the class with the highest probability. All padded regions were removed before calculating performance measures. Performance was evaluated using Cohen's kappa $\kappa_{p}$ per patient and overall accuracy $Acc$ and Cohen's kappa $\kappa_{c}$ . The final scores reported are the median  $\kappa_{p}$ and the overall $Acc$ and $\kappa_{c}$ (Table \ref{tb:summary_table_results}). The performance measures are also reported over the overall confusion matrix representing the prediction summary of the 30-seconds windows of a given test dataset (Figure \ref{fig:cf}). Finally, the performance measures were computed for the following tasks: four-class (wake, light, deep, REM), three-class (wake, REM, NREM) and two-class (wake, sleep) classifications. The significance of the results was calculated using a Wilcoxon signed rank test with a p-value significance threshold set to 0.05.

\subsubsection{Sleep measures}
In order to provide a better understanding of how these performances translate to support of downstream diagnosis tasks, the following sleep measures were evaluated: total sleep time (TST), sleep efficiency (SE), and fractions of various sleep stages (FRLight, FRDeep, FRREM). The TST measure is instrumental in revealing the model's ability to accurately track sleep duration. The SE measure helps in understanding how well the patient slept and plays a role in monitoring sleep quality. The fraction stages are important to determine whether the patient cycled through all stages during the night and at a sufficient percentage. A deficiency in a specific fraction stage suggest the possible presence of a sleep disorder. Calculating these measures and comparing them to the ground truth enhance our understanding of the clinical usability of the model. We report the mean absolute error (MAE) between the estimated and reference sleep measures for the SleepPPG-Net2 model over the grouped test sets. Sleep measures are defined as:

\begin{equation}
TST = \sum_{i=1}^{n} Light[i] + \sum_{i=1}^{n} Deep[i] + \sum_{i=1}^{n} REM[i]
\end{equation}

\begin{equation}
SE = \frac{TST}{TST + \sum_{1}^{n} Wake } \cdot 100
\end{equation}

\begin{equation}
FR{Stage} = \frac{ \sum_{1}^{n} Stage}{TST} \cdot 100
\end{equation}

where, assuming $n$ windows in a recording, the variables Wake, Light, Deep, and REM are binary arrays $\in\mathbb{R}^{n}$, encoding the presence or absence of respective sleep stages. The variable Stage $\in\mathbb{R}^{n}$ refers to the binary array representing the stage for which we are computing the sleep fraction. $Stage[i]$ represents the binary value in the i-th window of the Stage array, encoding whether the window is annotated as Stage or not.

\subsection{Error analysis}
Multivariate linear mixed-effects models were used to evaluate the effects of patient age, sex, AHI, BMI, ethnicity, as well as their interactions on the per-patient kappa performance measure. This analysis was performed on the MESA, CFS, ABC and HomePAP datasets since they had all the relevant metadata. The resulting standardized model coefficients are reported. In addition, for the CAP dataset the performance was displayed as a function of the sleep condition.

\begin{figure*}[b!]
  \centering
    \includegraphics[width=\textwidth]{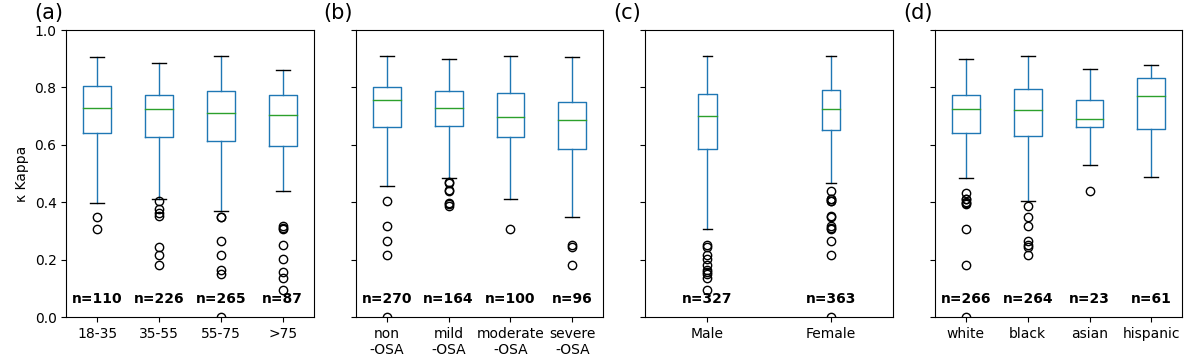}
    \caption{Boxplots representing the kappa per patient $\kappa_{p}$ by the (a) age, (b) OSA severity, (c) sex and (d) ethnicity. n: the number of patients. }
    \label{fig:mixed}
\end{figure*}
\section{Results}

\subsection{Performance statistics}
For the four-class classification task, the  SleepPPG-Net2 performance on the source domain test set, MESA-test, was the highest, with a kappa of 0.75 (0.67-0.81) compared to the SleepPPG-Net with a kappa of 0.74 (0.66-0.80) and BM-DTS approach, with a kappa of 0.65 (0.56-0.74) (Figures \ref{fig:res},\ref{tb:summary_table_results}). When using the SleepPPG-Net2 multisource domain training approach, the kappa on the target domains was 1-19\% higher compared to SleepPPG-Net (Table \ref{tb:summary_table_results}). The results were consistent for the three and two class classification tasks, with SleepPPG-Net2 outperforming the two benchmarks (Figures \ref{tb:summary_table_results},  Table S1 and Table S2).

\begin{table}[b!]
  \centering
  \renewcommand{\arraystretch}{1.2}
  \begin{tabular}{p{1.9cm} |p{0.9cm} p{0.9cm} p{0.9cm} p{0.9cm} p{0.9cm}}
    \hline
    {\textbf{Model / $MAE$}} 
    & \textbf{TST (min)} & \textbf{SE \newline (\%)} & \textbf{FRLight (\%)} & \textbf{FRDeep (\%)} & \textbf{FRREM (\%)}\\

    \hline 
    BM-DTS  & 43.48& 8.02& 13.05& 11.61& 4.73  \\ 
    SleepPPG-Net  & 27.64& 5.23& 10.05& 9.07& 3.87\\ 
    SleepPPG-Net2  & \textbf{24.2} &\textbf{4.56}&\textbf{9.47}&\textbf{8.53}&\textbf{3.72} \\
    \hline
  \end{tabular}
  \caption{Performance in estimating sleep measures. Presented are the mean absolute errors (MAE) of the six test sets. For each sleep measure, the least MAE is highlighted in bold.}
  \label{tb:summary_table_metrics}
\end{table}
\subsection{Sleep measures}
Results of assessment of SleepPPG-Net2 ability to estimate standard sleep measures are presented in Figure \ref{fig:correlation} and Table \ref{tb:summary_table_metrics}. Some sleep measures could be estimated with a good approximation (TST (MAE=24.2 min), SE (MAE=4.56\%), FRREM and FRNREM  (MAE=3.72\%)). Its performance in estimating FRDeep (MAE=8.53\%) and FRLight (MAE=9.47\%) were poorer.

\subsection{Error analysis}
The analysis from the mixed linear regression model revealed that age and AHI had statistically significant effects on the patient's kappa score $\kappa_{p}$ with a $p-value$ of less than $0.01$. Specifically, the model showed that as age increased, $\kappa_{p}$ decreased with a coefficient of $-0.216$, suggesting that younger individuals tend to have higher $\kappa_{p}$ scores. Likewise, a rise in AHI corresponded to a decrease in $\kappa_{p}$ with a coefficient of $-0.175$, highlighting the negative impact of OSA severity on kappa performance.  Sex was also identified as a significant factor, with a $p-value$ of less than $0.05$ and a coefficient of $-0.205$, indicating that men have a lower $\kappa_{p}$ compared to women. The model showed no significant impact of ethnicity on $\kappa_{p}$. Figure \ref{fig:mixed} shows boxplots for performance across age, OSA, sex and ethnicity.

The error analysis for the relatively low kappa in the CAP dataset shows that the RBD sleep disorder seems to impact the kappa performance (Figure S1 (a)). Indeed, when examining the confusion matrix for the 16 RBD patients, as shown in Figure S1 (b), the REM stage is frequently misclassified as the Wake stage (44\%) whereas it was systematically under 3\% for the other datasets (Figure \ref{fig:cf}).



\section{Discussion and conclusion}
In this research, we developed and evaluated a deep learning model, denoted SleepPPG-Net2, which employs a multisource domain training approach to enable a more generalizable representation for sleep staging.  SleepPPG-Net2 improved between 1-19\% in $\kappa_{p}$ performance (Table \ref{tb:summary_table_results}, Figure \ref{fig:res}) over the best benchmark for the different target domain datasets. SleepPPG-Net2 was systematically higher than BM-DTS benchmark, showing that the use of raw PPG versus beat-to-beat interval variation is a valuable approach.  The evaluation of sleep measures was also significantly improved, with the lowest MAE obtained for SleepPPG-Net2 (Table \ref{tb:summary_table_metrics}, Figure \ref{fig:cf}). The present study demonstrated that the multi-source training approach significantly enhanced the performance of our original SleepPPG-Net model \cite{kotzen_sleepppg-net_2023}. Specifically, the MAEs obtained show that SleepPPG-Net2 is a viable algorithm to estimate TST, SE and FRREM. These measures can be used to support the diagnosis of insomnia, hypersomnia, sleep apnea and periodic limb movement disorder. In constrast, the MAEs obtained for estimating FRLight and FRDeep, although improved compared to previous models, were still too high. This may limit the usage of such a model in identifying disrupted sleep and very specific sleep disorders such as Disorders of Arousal parasomnias. 

The error analysis revealed that increased age, the OSA severity and male sex are independent factors leading to a decreased $\kappa_{p}$. OSA disrupts the normal sleep structure with frequent arousals, variations in heart rate and blood flow. This may make the extraction of sleep staging patterns in PPG more challenging. Aging is associated with changes that impact cardiovascular health, such as reduced vascular compliance and alterations in peripheral circulation. These changes may affect the PPG waveform, rendering sleep staging from PPG more challenging. Interestingly, we found that ethnicity did not significantly affect performance. This is in contradiction to several recent studies \cite{sinaki_ethnic_2022,levy_deep_2023} that demonstrated the effect of ethnicity on pulse oximetry reading performance. We note, however, that MESA, which was used for training, had a fair proportion of different ethnic groups (Figure \ref{fig:distribution_violin} (d)).
The error analysis conducted on the CAP dataset leads us to hypothesize that abnormalities in the Autonomic Nervous System (ANS), which are present in RBD, might cause changes in the PPG signal. 

This study was the first to evaluate the generalizability of deep learning algorithms for the task of sleep staging from PPG time-series. The study had, however, some limitations. It incorporated six open datasets that cover a diverse range of population samples. However, several comorbidities common in populations referred for a sleep analysis were either not present or underrepresented. This includes individuals with atrial fibrillation or other arrhythmias, which could potentially make the sleep staging task more difficult. Additionally, the datasets scarcely include individuals with central sleep apnea and other less common sleep disorders, such as narcolepsy, REM sleep behavior disorder, or periodic limb movement disorder. Further research is needed to validate our methodology in these specific patient groups. Finally, PPG was recorded mostly or exclusively with Nonin fingertip oximeters. Whether the model is robust to distribution shifts relative to the oximeter manufacturer and to the probe location (e.g. smart rings which sense PPG from the base of the finger) remains to be determined.


In conclusion, SleepPPG-Net2 establishes a new state-of-the-art for sleep staging from PPG. The good performance of SleepPPG-Net2 in estimating various common sleep measures, combined with the widespread adoption of wearable sensors that measure PPG, opens up numerous possibilities for medical applications.

\section*{References}
\vspace*{-0.7cm}
\bibliographystyle{IEEEtran}
\bibliography{zotero, IEEEabrv}

\section{Acknowledgement} 
This research was supported by funding from the Israel Innovation Authority. The research was supported by a cloud computing grant from the Israel Council of Higher Education, administered by the Israel Data Science Initiative. We acknowledge the assistance of ChatGPT, an AI-based language model developed by OpenAI, for its help in editing the English language of this manuscript. 

MESA was funded by NIH-NHLBI Association of Sleep Disorders with Cardiovascular Health Across Ethnic Groups (RO1 HL098433). MESA is supported by NHLBI funded contracts HHSN268201500003I, N01-HC-95159, N01-HC-95160, N01-HC-95161, N01-HC-95162, N01-HC-95163, N01-HC-95164, N01-HC-95165, N01-HC-95166, N01-HC-95167, N01-HC-95168 and N01-HC-95169 from the National Heart, Lung, and Blood Institute, and by cooperative agreements UL1-TR-000040, UL1-TR-001079, and UL1-TR-001420 funded by NCATS. CFS was supported by grants from the National Institutes of Health (HL46380, M01 RR00080-39, T32-HL07567, RO1-46380). ABC was supported by National Institutes of Health grants R01HL106410 and K24HL127307. Philips Respironics donated the CPAP machines and supplies used in the perioperative period for patients undergoing bariatric surgery. HomePAP was supported by the American Sleep Medicine Foundation 38-PM-07 Grant: Portable Monitoring for the Diagnosis and Management of OSA. The National Sleep Research Resource was supported by the National Heart, Lung, and Blood Institute (R24 HL114473, 75N92019R002).

\end{document}


\renewcommand{\figurename}{Figure S}
\renewcommand{\tablename}{Table S}

\onecolumn
\section{Supplementary figures}

\begin{figure}[h]
  \centering
    \includegraphics[width=\textwidth]{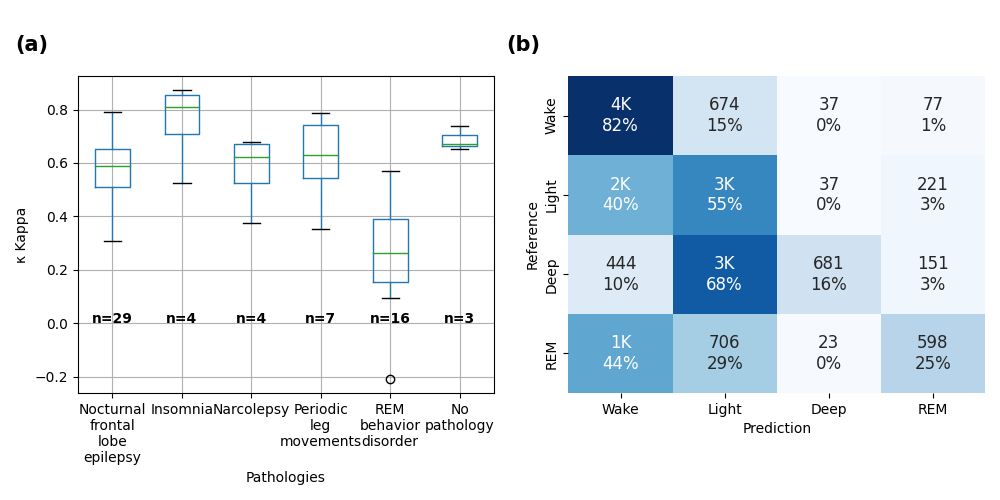}
    \caption{Performances on diseases in CAP: Panel (a) represents a boxplot of the kappa $\kappa_{p}$ (four-class) for each sleep disorder present in CAP. Panel (b) is a confusion matrix (four-class) of the 16 CAP patients with RBD. }
    \label{fig:cap_error_analysis}
\end{figure}

\FloatBarrier

\section{Supplementary tables}

\begin{table*}[h]
  \centering
  \renewcommand{\arraystretch}{1.2}
  \begin{tabular}{p{1.8cm} |p{0.4cm} p{0.4cm} p{0.4cm} | p{0.4cm} p{0.4cm} p{0.4cm}  |p{0.4cm} p{0.4cm} p{0.4cm} | p{0.4cm} p{0.3cm} p{0.4cm} | p{0.4cm} p{0.4cm} p{0.4cm} | p{0.4cm} p{0.4cm} p{0.4cm}}
    \hline
    \multirow{3}{3cm}
    {\textbf{Model}} 
    & \multicolumn{18}{c}{\textbf{datasets}} \\
    \cline{2-19} 
    & \multicolumn{3}{c}{\textbf{MESA}} 
    & \multicolumn{3}{c}{\textbf{CFS}} 
    & \multicolumn{3}{c}{\textbf{ABC}} 
    & \multicolumn{3}{c}{\textbf{HomePAP}} 
    & \multicolumn{3}{c}{\textbf{CAP}} 
    & \multicolumn{3}{c}{\textbf{SLEEPAI}}\\ 
    \cline{2-19} 
    & \textbf{$\kappa_{p}$} & $\kappa_{c}$ & \textbf{$Acc$} & \textbf{$\kappa_{p}$} & $\kappa_{c}$ &  \textbf{$Acc$} & \textbf{$\kappa_{p}$} & $\kappa_{c}$ &  \textbf{$Acc$} & \textbf{$\kappa_{p}$} & $\kappa_{c}$ &  \textbf{$Acc$} & \textbf{$\kappa_{p}$} & $\kappa_{c}$ &  \textbf{$Acc$}& \textbf{$\kappa_{p}$} & $\kappa_{c}$ &  \textbf{$Acc$}\\
    \hline 
    DTS  & 0.73 & 0.72 & 0.83  & 0.72 & 0.70 & 0.82  & 0.66 & 0.66 & 0.82  & 0.62 & 0.58 & 0.77   & 0.57 & 0.57 & 0.76 & 0.63 & 0.64 & 0.81  \\ 
    SleepPPG-Net  & 0.82 & \textbf{0.82} & \textbf{0.89} & 0.82 & 0.79 & \textbf{0.88} & 0.74 & 0.73 & 0.85  & 0.72 & 0.72 & 0.84 & 0.66 & 0.64 & 0.80 & 0.73 & 0.70 & 0.85\\ 
    SleepPPG-Net2  & \textbf{0.83} & \textbf{0.82} & \textbf{0.89} & \textbf{0.83} & \textbf{0.80} & \textbf{0.88} & \textbf{0.75} & \textbf{0.74} & \textbf{0.86} &  \textbf{0.77} & \textbf{0.76} & \textbf{0.86} &  \textbf{0.72} & \textbf{0.66} & \textbf{0.82} &  \textbf{0.74} & \textbf{0.72} & \textbf{0.86} \\
    \hline
  \end{tabular}
  \caption{Summary table presenting the median kappa, overall kappa and accuracy for the benchmark models (DTS, SleepPPG-Net) and the model developed in this research (SleepPPG-Net2) on the six datasets for three-class sleep staging. MESA, source domain test set. The other datasets are target domains. }
  \label{tb:3C}
\end{table*}


\begin{table*}[h]
  \centering
  \renewcommand{\arraystretch}{1.2}
  \begin{tabular}{p{1.8cm} |p{0.4cm} p{0.4cm} p{0.4cm} | p{0.4cm} p{0.4cm} p{0.4cm}  |p{0.4cm} p{0.4cm} p{0.4cm} | p{0.4cm} p{0.3cm} p{0.4cm} | p{0.4cm} p{0.4cm} p{0.4cm} | p{0.4cm} p{0.4cm} p{0.4cm}}
    \hline
    \multirow{3}{3cm}
    {\textbf{Model}} 
    & \multicolumn{18}{c}{\textbf{datasets}} \\
    \cline{2-19} 
    & \multicolumn{3}{c}{\textbf{MESA}} 
    & \multicolumn{3}{c}{\textbf{CFS}} 
    & \multicolumn{3}{c}{\textbf{ABC}} 
    & \multicolumn{3}{c}{\textbf{HomePAP}} 
    & \multicolumn{3}{c}{\textbf{CAP}} 
    & \multicolumn{3}{c}{\textbf{SLEEPAI}}\\ 
    \cline{2-19} 
    & \textbf{$\kappa_{p}$} & $\kappa_{c}$ & \textbf{$Acc$} & \textbf{$\kappa_{p}$} & $\kappa_{c}$ &  \textbf{$Acc$} & \textbf{$\kappa_{p}$} & $\kappa_{c}$ &  \textbf{$Acc$} & \textbf{$\kappa_{p}$} & $\kappa_{c}$ &  \textbf{$Acc$} & \textbf{$\kappa_{p}$} & $\kappa_{c}$ &  \textbf{$Acc$}& \textbf{$\kappa_{p}$} & $\kappa_{c}$ &  \textbf{$Acc$}\\
    \hline 
    DTS  & 0.76 & 0.74 & 0.88  & 0.74 & 0.71 & 0.87  & 0.59 & 0.63 & 0.86 & 0.62 & 0.56 & 0.82  & 0.52 & 0.55 & 0.83 & 0.61 & 0.64 & 0.87  \\ 
    SleepPPG-Net  & 0.86 & 0.85 & 0.93 & 0.84 & 0.82 & 0.91 & 0.68 & 0.71 & 0.89  & 0.70 & 0.71 & 0.89 & 0.55 & 0.60 & 0.85 & \textbf{0.74} & 0.72 & 0.90\\ 
    SleepPPG-Net2  & \textbf{0.87} & \textbf{0.86} & \textbf{0.93} & \textbf{0.86} & \textbf{0.83} & \textbf{0.92} & \textbf{0.74} & \textbf{0.72} & \textbf{0.89} &  \textbf{0.77} & \textbf{0.75} & \textbf{0.90} &  \textbf{0.67} & \textbf{0.64} & \textbf{0.87} &  0.73 & \textbf{0.73} & \textbf{0.91} \\
    \hline
  \end{tabular}
  \caption{Summary table presenting the median kappa, overall kappa and accuracy for the benchmark models (DTS, SleepPPG-Net) and the model developed in this research (SleepPPG-Net2) on the six datasets for two-class sleep staging. MESA, source domain test set. The other datasets are target domains.}
  \label{tb:2C}
\end{table*}